\documentclass[10pt, a4paper]{article}
\usepackage{lrec}
\usepackage{inputenc}
\usepackage{graphicx}
\usepackage{tabularx}
\usepackage{soul}
\usepackage{enumitem}
\usepackage{linguex}
\usepackage{booktabs}
\usepackage{amsmath, amssymb, mathtools, bm}

\usepackage{epstopdf}

\usepackage{hyperref}
\usepackage{xstring}

\title{The Universal Decompositional Semantics Dataset and Decomp Toolkit}

\name{Aaron Steven White$^\triangleright$, Elias Stengel-Eskin$^\triangleleft$, Siddharth Vashishtha$^\triangleright$,\\\textbf{\large Venkata Govindarajan$^\ddagger$, Dee Ann Reisinger$^\triangleleft$, Tim Vieira$^\triangleleft$, Keisuke Sakaguchi$^\dagger$,}\\\textbf{\large Sheng Zhang$^\triangleleft$, Francis Ferraro$^\circ$, Rachel Rudinger$^\dagger$, Kyle Rawlins$^\triangleleft$, Benjamin Van Durme$^\triangleleft$}}

\address{$^\triangleright$University of Rochester, $^\triangleleft$Johns Hopkins University, $^\circ$University of Maryland Baltimore County\\$^\dagger$Allen Institute for Artificial Intelligence, $^\ddagger$University of Texas at Austin \\}

\abstract{
We present the Universal Decompositional Semantics (UDS) dataset (v1.0), which is bundled with the Decomp toolkit (v0.1). UDS1.0 unifies five high-quality, decompositional semantics-aligned annotation sets within a single semantic graph specification---with graph structures defined by the predicative patterns produced by the PredPatt tool and real-valued node and edge attributes constructed using sophisticated normalization procedures. The Decomp toolkit provides a suite of Python 3 tools for querying UDS graphs using SPARQL. Both UDS1.0 and Decomp0.1 are publicly available at \url{http://decomp.io}. \\ \newline  }

\begin{document}

\setlength{\Exlabelsep}{0em}%
\setlength{\SubExleftmargin}{1.2em}%
\setlength{\Extopsep}{.3\baselineskip}%

\maketitleabstract

\section{Introduction}

Traditional semantic annotation frameworks generally define complex, often exclusive category systems that require highly trained annotators to build \cite{palmer_proposition_2005,banarescu_abstract_2013,abend_universal_2013,oepen_semeval_2014,oepen_semeval_2015,bos_groningen_2017,abzianidze_parallel_2017,abzianidze_towards_2017,schneider_comprehensive_2018}. And in spite of their high quality for the cases they are designed to handle, these frameworks can be brittle to cases that (i) deviate from prototypical instances of a category; (ii) are equally good instances of multiple categories; or (iii) fall under a category that was erroneously excluded from the framework's ontology.\footnote{\newcite{shalev_preparing_2019} discuss multiple recent, instructive examples of such brittleness.} 

An alternative approach to semantic annotation that addresses these issues has been growing in popularity: decompositional semantics \cite{reisinger_semantic_2015,white_universal_2016}. In this approach, which is rooted in a long tradition of theoretical approaches to lexical semantics \cite[and references therein]{pustejovsky_generative_1995,levin_argument_2005}, semantic annotation takes the form of many simple questions about words or phrases (in context) that are easy for na\"ive native speakers to answer, thus allowing annotations to be crowd-sourced while retaining high interannotator agreement. 

The decompositional approach can be thought of as a feature-based counterpart to traditional category-based systems, with each question determining a semantic feature. Common feature configurations often correspond to categories in a traditional framework \cite{reisinger_semantic_2015,govindarajan_decomposing_2019}; but unlike such frameworks, a decompositional approach retains the ability to capture configurations that were not considered at design time. Further, unlike a categorical framework, reannotation after an overhaul of the framework's ontology is never necessary, since additional annotations simply accrue to sharpen the framework's ability to capture fine-grained semantic phenomena. 

A variety of semantic annotation datasets that take a decompositional approach now exist, including ones that target semantic roles \cite{reisinger_semantic_2015,white_universal_2016}, entity types \cite{white_universal_2016}, event factuality \cite{white_universal_2016,rudinger_neural_2018}, linguistic expressions of generalizations about entities and events \cite{govindarajan_decomposing_2019}, and temporal properties of and relations between events \cite{vashishtha_fine-grained_2019}. But despite the potential benefits of a decompositional approach---as well as the broad coverage of linguistic phenomena it has been shown to afford---a remaining obstacle to widespread adoption is the lack of a unified interface to these resources: prior work in UDS has approached annotation piecemeal---each effort focused on a restricted set of linguistic phenomena---without a broader push toward creating a unified semantic parsing resource.


To remedy this situation, we present the Universal Decompositional Semantics (UDS) dataset (v1.0) and the Decomp toolkit (v0.1), which we make publicly available at \url{http://decomp.io}. UDS1.0 unifies the five high-quality, decompositional semantics-aligned annotation sets listed above within a single semantic graph specification---with graph structures defined by the predicative patterns produced by the PredPatt tool and real-valued node and edge attributes constructed using sophisticated response normalization procedures. The Decomp toolkit provides a suite of Python 3 tools that make working with these data seamless, enabling a wide range of queries on Universal Decompositional Semantics graphs using the SPARQL 1.1 query language.

\begin{figure*}
    \centering
    \includegraphics[width=\textwidth]{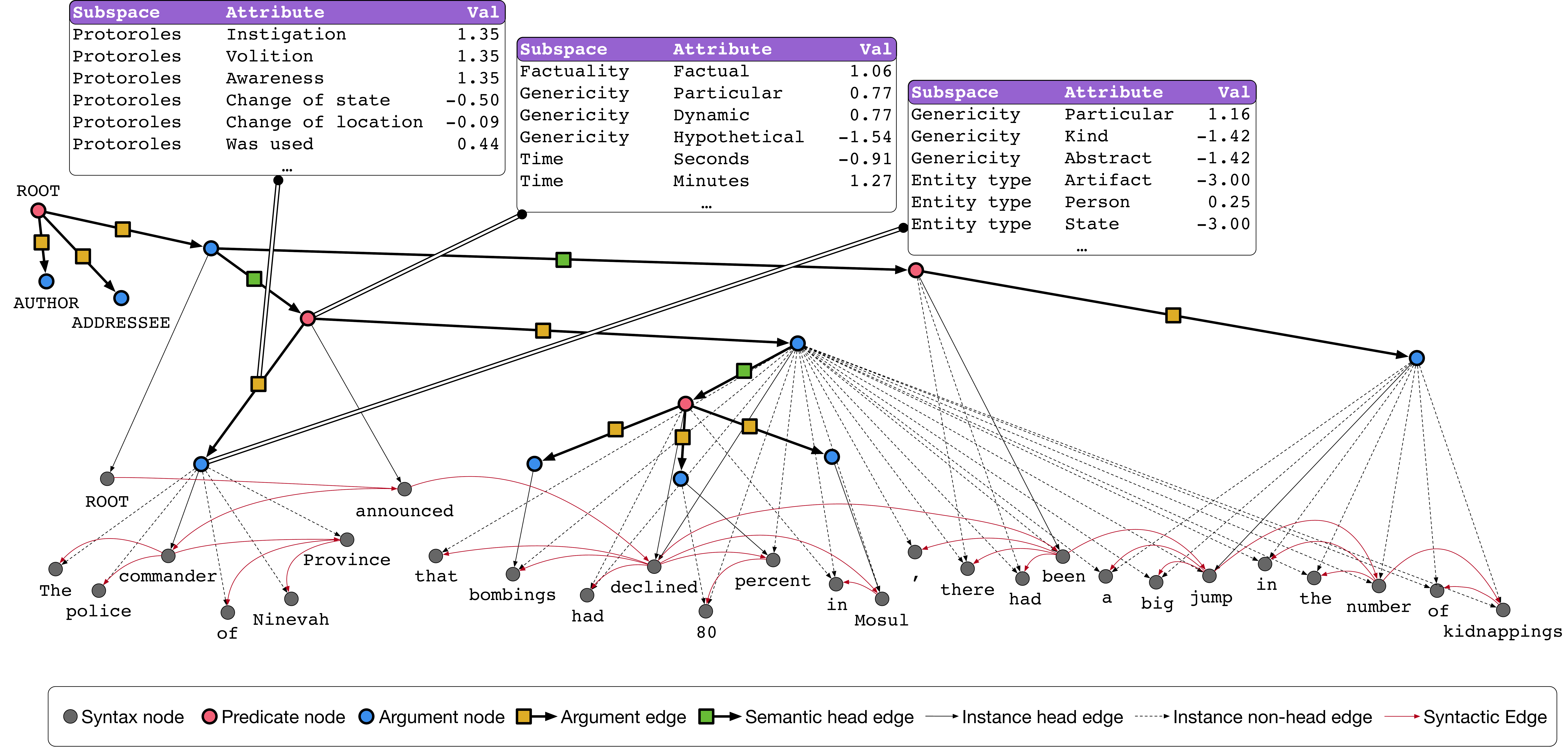}
    \caption{An example Universal Decompositional Semantics graph. Some semantic type information and most syntactic structure information (e.g. dependency relation and part-of-speech tags) are not shown but are available in the dataset.}
    \label{fig:udsgraph}
\end{figure*}

\section{Data}

UDS1.0 consists of three layers of annotations built on top of the English Web Treebank~\cite[EWT]{bies_english_2012}: (i) syntactic graphs (\S\ref{sec:syntacticgraphs}) built from existing gold Universal Dependencies (UD) parses on EWT \cite{nivre_universal_2015}; (ii) semantic graphs (\S\ref{sec:semanticgraphs}) built from the predicate-argument structures deterministically extracted from those parses using the PredPatt tool \cite{white_universal_2016,zhang_evaluation_2017}; and (iii) semantic types (\S\ref{sec:semantictypes}) for the predicates, arguments, and their relationships, derived from five decompositional semantics-aligned datasets. Figure \ref{fig:udsgraph} shows an example UDS graph with all three layers of annotation.

\subsection{Syntactic Graph}
\label{sec:syntacticgraphs}

The syntactic graphs that form the first layer of annotation in the dataset come from gold UD dependency parses provided in the UD-EWT treebank (v1.2), which contains sentences from the Linguistic Data Consortium's constituency parsed EWT in CoNLL-U format. UDS1.0 inherits UD-EWT's training, development, and test splits.

In UDS1.0, each dependency parsed sentence in UD-EWT is represented as a rooted directed graph (digraph). Each token in a sentence is associated with a node that has, at minimum, the following attributes:
\vspace{-1mm}
\begin{itemize}[itemsep=-3pt]
    \item \verb+position+ (\verb+int+): the ordinal position of that node as an integer (1-indexed)
    \item \verb+domain+ (\verb+str+): the subgraph this node is part of (always \verb+syntax+)
    \item \verb+type+ (\verb+str+): the type of the object in the particular domain (always \verb+token+)
    \item \verb+form+ (\verb+str+): the actual token
    \item \verb+lemma+ (\verb+str+): the lemma corresponding to the token
    \item \verb+upos+ (\verb+str+): the UD part-of-speech tag
    \item \verb+xpos+ (\verb+str+): the Penn TreeBank part-of-speech tag
\end{itemize}
\vspace{-1mm}
In addition, any attribute found in the UD features column of the CoNLL-U are inherited as node attributes by the syntactic graph.

Each graph also has a special root node that always has \texttt{domain} and \texttt{type} attributes set to \texttt{root}. Edges within the graph represent the grammatical relations annotated in UD-EWT and are directed, always pointing from the head to the dependent of the relation. At minimum, each edge has the following attributes:
\vspace{-1mm}
\begin{itemize}[itemsep=-3pt]
    \item \verb+domain+ (\verb+str+): the subgraph this node is part of (always \verb+syntax+)
    \item \verb+type+ (\verb+str+): the type of the object in the particular domain (always \verb+dependency+)
    \item \verb+deprel+ (\verb+str+): the UD dependency relation tag
\end{itemize}
\vspace{-1mm}
For information about the values \texttt{upos}, \texttt{xpos}, \texttt{deprel}, and the attributes contained in the features column can take on, see the \href{https://universaldependencies.org/guidelines.html}{UD Guidelines}.

\subsection{Semantic Graphs}
\label{sec:semanticgraphs}

The semantic graphs that form the second layer of annotation in the dataset are produced by the PredPatt system \cite{white_universal_2016,zhang_evaluation_2017}. PredPatt takes as input a UD parse and produces a set of predicates and set of arguments of each predicate. Both predicates and arguments are associated with a single head token in the sentence as well as a set of tokens that make up the predicate or argument (its span). Predicate or argument spans may be trivial in only containing the head token.

For example, given the dependency parse for the sentence \Next and its UD parse, PredPatt produces the following.

\ex. Chris$_1$ gave$_2$ the$_3$ book$_4$ to$_5$ Pat$_6$

\vspace{-3mm}

\begin{verbatim}
   ?a gave ?b to ?c
      ?a: Chris
      ?b: the book
      ?c: Pat
\end{verbatim}
\vspace{-1mm}    
Assuming UD's 1-indexation, the single predicate in this sentence (\textit{gave...to}) has a head at position 2 and a span over positions \{2, 5\}. This predicate has three arguments, one headed by \textit{Chris} at position 1, with span over position \{1\}; one headed by \textit{book} at position 4, with span over positions \{3, 4\}; and one headed by \textit{Pat} at position 6, with span over position \{6\}.\footnote{See the \href{https://github.com/hltcoe/PredPatt/blob/master/doc/DOCTEST.md}{PredPatt documentation tests} for further examples.}

Each predicate and argument produced by PredPatt is associated with a node in a digraph. At minimum, each such node has the following attributes:
\vspace{-1mm}
\begin{itemize}[itemsep=-3pt]
    \item \verb+domain+ (\verb+str+): the subgraph this node is part of (always semantics)
    \item \verb+type+ (\verb+str+): the type of the object in the particular domain (either \verb+predicate+ or \verb+argument+)
\end{itemize}
\vspace{-1mm}
Predicate and argument nodes produced by PredPatt furthermore always have at least one outgoing instance edge that points to nodes in the syntax domain that correspond to the associated span of the predicate or argument. At minimum, each such edge has the following attributes.
\vspace{-1mm}
\begin{itemize}[itemsep=-3pt]
    \item \verb+domain+ (\verb+str+): the subgraph this node is part of (always \verb+interface+)
    \item \verb+type+ (\verb+str+): the type of the object in the particular domain (either \verb+head+ or \verb+nonhead+)
\end{itemize}
\vspace{-1mm}
Because PredPatt produces a unique head for each predicate and argument, there is always exactly one instance edge of type head from any particular node in the semantics domain. There may or may not be instance edges of type nonhead.

In addition to instance edges, predicate nodes always have exactly one outgoing edge connecting them to each of the nodes corresponding to their arguments. At minimum, each such edge has the following attributes.
\vspace{-1mm}
\begin{itemize}[itemsep=-3pt]
    \item \verb+domain+ (\verb+str+): the subgraph this node is part of (always \verb+semantics+)
    \item \verb+type+ (\verb+str+): the type of the object in the particular domain (always \verb+dependency+)
\end{itemize}
\vspace{-1mm}
There is one special case where an argument node has an outgoing edge that points to a predicate node: clausal subordination. For example, given the dependency parse for the sentence \Next, PredPatt produces the following:

\ex. Gene$_1$ thought$_2$ that$_3$ Chris$_4$ gave$_5$ the$_6$ book$_7$ to$_8$ Pat$_9$

\vspace{-3mm}

\begin{verbatim}
   ?a thinks ?b
      ?a: Gene
      ?b: SOMETHING := that Chris gave 
                       the book to Pat

   ?a gave ?b to ?c
      ?a: Chris
      ?b: the book
      ?c: Pat
\end{verbatim}
\vspace{-1mm}
In this case, the second argument of the predicate headed by \textit{thinks} is the argument that \textit{Chris gave the book to Pat}, which is headed by \textit{gave}. This argument is associated with a node of type argument with span over positions \{3, 4, 5, 6, 7, 8, 9\}. In addition, there is a predicate headed by \textit{gave}. This predicate is associated with a node with span over positions \{5, 8\}. This predicate node in turn has an outgoing edge pointing to the argument node. At minimum, each such edge has the following attributes:
\vspace{-1mm}
\begin{itemize}[itemsep=-3pt]
    \item \verb+domain+ (\verb+str+): the subgraph this node is part of (always \verb+semantics+)
    \item \verb+type+ (\verb+str+): the type of the object in the particular domain (always \verb+head+)
\end{itemize}
\vspace{-1mm}
The type attribute in this case has the same value as instance edges (\verb+head+), but crucially the domain attribute is distinct. In the case of instance edges, it is \verb+interface+ and in the case of clausal subordination, it is \verb+semantics+. This matters when making queries against the graph and serializing to an RDF-based format.

Every semantic graph contains at least four additional \textit{performative} nodes that are not produced by PredPatt.\footnote{The term \textit{performative} because these nodes are intended to represent something akin to analogous syntactically represented nodes argued for by \newcite{ross_act_1972}.}
\vspace{-1mm}
\begin{itemize}[itemsep=-3pt]
    \item an argument node representing the entire sentence in the same way complement clauses are represented
    \item a predicate node representing the author’s production of the entire sentence directed at the addressee
    \item an argument node representing the author
    \item an argument node representing the addressee
\end{itemize}
\vspace{-1mm}
All of these nodes have a \verb+domain+ attribute with value \verb+semantics+. Unlike nodes associated with PredPatt predicates and arguments, the predicate node representing the author’s production, the argument node representing the author, and the argument node representing the addressee.  The argument node representing the entire sentence does, however, have an instance head edge to the syntactic root node. This node also has semantics head edges to each of the predicate nodes in the graph that are not dominated by any other semantics node. The predicate node representing the author's production in turn has an argument edge to each of the three argument nodes listed above.

These performative nodes are included for purposes of forward compatibility. None of them currently have attributes, but future releases of decomp will include annotations on either them or their edges.

\begin{table}[t]
\footnotesize
    \centering
{\large\bf Annotated Nodes}\\\vspace{1mm}
{\normalsize\it Train}\\\vspace{1mm}
\begin{tabular}{lrrrr}
\toprule
{} &  Factuality &  Genericity &   Time &  Entity Type \\
\midrule
Factuality  &       21,092 &       20,929 &  20,733 &            0 \\
Genericity  &       {   } &       56,594 &  26,314 &        16,873 \\
Time        &       {   } &       {   } &  26,324 &            0 \\
Entity Type &       {   } &       {   } &  {   } &        17,192 \\
\bottomrule
\end{tabular}

\vspace{3mm}{\normalsize\it Dev}\\\vspace{1mm}
\begin{tabular}{lrrrr}
\toprule
{} &  Factuality &  Genericity &  Time &  Entity Type \\
\midrule
Factuality  &        2,476 &        2,456 &  2,320 &            0 \\
Genericity  &        {  } &        6,858 &  3,051 &         1,894 \\
Time        &        {  } &        {  } &  3,051 &            0 \\
Entity Type &        {  } &        {  } &  {  } &         1,943 \\
\bottomrule
\end{tabular}

\vspace{3mm}{\normalsize\it Test}\\\vspace{1mm}
\begin{tabular}{lrrrr}
\toprule
{} &  Factuality &  Genericity &  Time &  Entity Type \\
\midrule
Factuality  &        2,413 &        2,394 &  2,275 &            0 \\
Genericity  &        {  } &        6,602 &  2,927 &         1,847 \\
Time        &        {  } &        {  } &  2,927 &            0 \\
Entity Type &        {  } &        {  } &  {  } &         1,876 \\
\bottomrule
\end{tabular}

    \caption{The number of predicate and argument nodes that are annotated for both the node type subspace along the columns and the one along the rows. The diagonal elements show the total number of nodes annotated for a particular node type subspace. The Entity Type subspace is not annotated on any of the same nodes as Factuality and Time because Entity Type is only annotated on arguments and Factuality and Time are only annotated on predicates.}
    \label{tab:nodextab}
\end{table}

\subsection{Semantic Types}
\label{sec:semantictypes}

PredPatt makes very coarse-grained typing distinctions---between predicate and argument nodes, on the one hand, and between dependency and head edges, on the other. UDS provides ultra fine-grained typing distinctions, represented as collections of real-valued attributes. The collection of all node and edge attributes defined in UDS determines the \textit{UDS type space}; any cohesive subset determines a \textit{UDS type subspace}.\footnote{It is important to note that, though all nodes and edges in the semantics domain have a type attribute, UDS does not afford any special status to these types. That is, the only thing that UDS ``sees'' are the nodes and edges in the semantics domain. This implies that the set of nodes and edges visible to UDS is, in principle, a superset of those associated with PredPatt predicates and their arguments. For UDS1.0, however, we only include attributes of nodes and edges built from predicate-argument patterns produced by PredPatt. In future releases, additional edge types will be added. For instance, predicate-predicate temporal relations are currently annotated in the temporal relations dataset from which we extract temporal durations \cite{vashishtha_fine-grained_2019}.}
 
\subsubsection{Deriving attribute values and confidence scores}

UDS attributes are derived from crowd-sourced annotations of the heads or spans corresponding to predicates and/or arguments (described in \S\ref{sec:nodeattributes} and \S\ref{sec:edgeattributes}) and are represented in the dataset as node or edge attributes. All of these attributes come from existing datasets that are publicly available at \url{http://decomp.io}.\footnote{All of these datasets are introduced in peer-reviewed publications which report extensive interannotator agreement statistics, and so we do not report such statistics here.} Table \ref{tab:nodextab} provides a breakdown of the number of PredPatt argument and predicate nodes annotated for different node type subspaces. In total, there are 57,080 annotated nodes in the training set, 6,927 in the development set, and 6,650 in the test set. Table \ref{tab:edgextab} provides a similar breakdown, showing the number of PredPatt edges that are annotated for an edge type subspace and touch nodes that are annotated for different node type subspaces. In total, there are 5,669 annotated edges in the training set, 751 in the development set, and 670 in the test set.  

\begin{table}[t]
\footnotesize
    \centering
{\large\bf Annotated Edges + Nodes}\\\vspace{1mm}
{\normalsize\it Train}\\\vspace{1mm}
\begin{tabular}{lrrrr}
\toprule
{} &  Factuality &  Genericity &  Time &  Entity Type \\
\midrule
Factuality  &           0 &        3,935 &     0 &         2,670 \\
Genericity  &        {  } &        4,200 &  4,163 &         2,903 \\
Time        &        {  } &        {  } &     0 &         2,883 \\
Entity Type &        {  } &        {  } &  {  } &            0 \\
\bottomrule
\end{tabular}

\vspace{3mm}{\normalsize\it Dev}\\\vspace{1mm}
\begin{tabular}{lrrrr}
\toprule
{} &  Factuality &  Genericity &  Time &  Entity Type \\
\midrule
Factuality  &           0 &         536 &     0 &          340 \\
Genericity  &         { } &         570 &   542 &          365 \\
Time        &         { } &         { } &     0 &          344 \\
Entity Type &         { } &         { } &   { } &            0 \\
\bottomrule
\end{tabular}

\vspace{3mm}{\normalsize\it Test}\\\vspace{1mm}
\begin{tabular}{lrrrr}
\toprule
{} &  Factuality &  Genericity &  Time &  Entity Type \\
\midrule
Factuality  &           0 &         481 &     0 &          298 \\
Genericity  &         { } &         507 &   471 &          320 \\
Time        &         { } &         { } &     0 &          296 \\
Entity Type &         { } &         { } &   { } &            0 \\
\bottomrule
\end{tabular}

    \caption{The number of edges that are annotated for semantic protoroles and which touch predicate and argument nodes, where one is annotated for at least the node type subspace along the columns and the other is annotated for the one along the rows. Zeros arise for node type subspaces that are only annotated for either predicate or argument nodes---e.g. because the Time subspace is only annotated on predicates, Time cannot be annotated for both nodes. (This will change in future versions of UDS.)}
    \label{tab:edgextab}
    \vspace{-5mm}
\end{table}

\begin{figure*}[t]
    \includegraphics[width=0.18\textwidth]{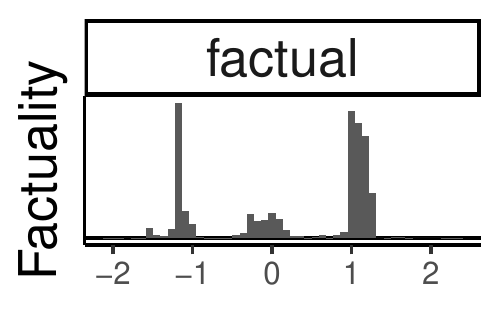}
    
    \includegraphics[width=\textwidth]{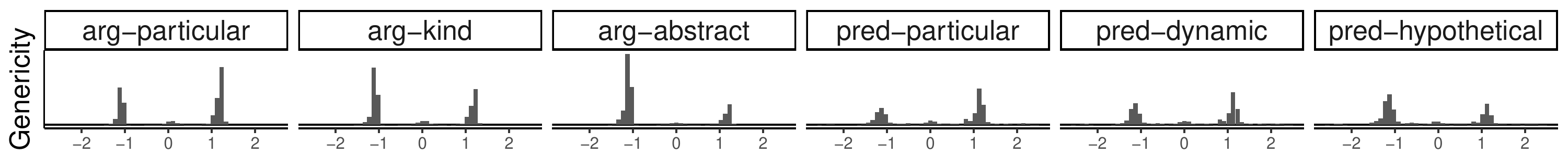}
    
    \includegraphics[width=\textwidth]{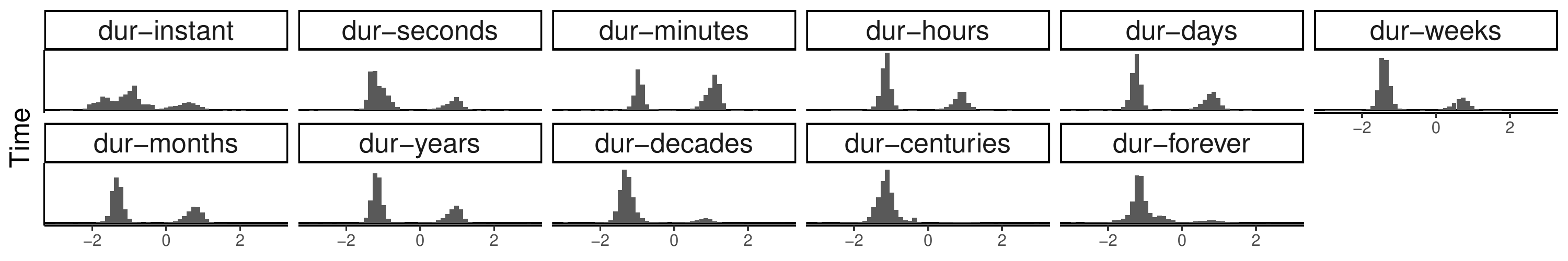}
    
    \includegraphics[width=\textwidth]{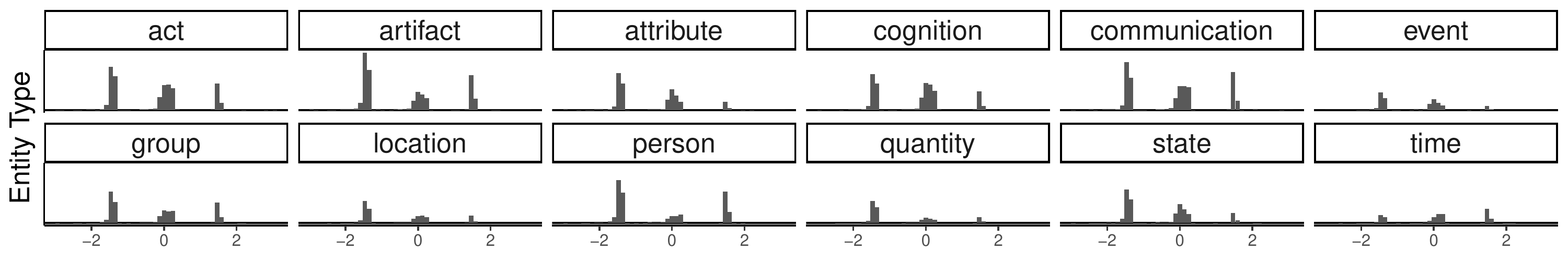}
    
    \includegraphics[width=\textwidth]{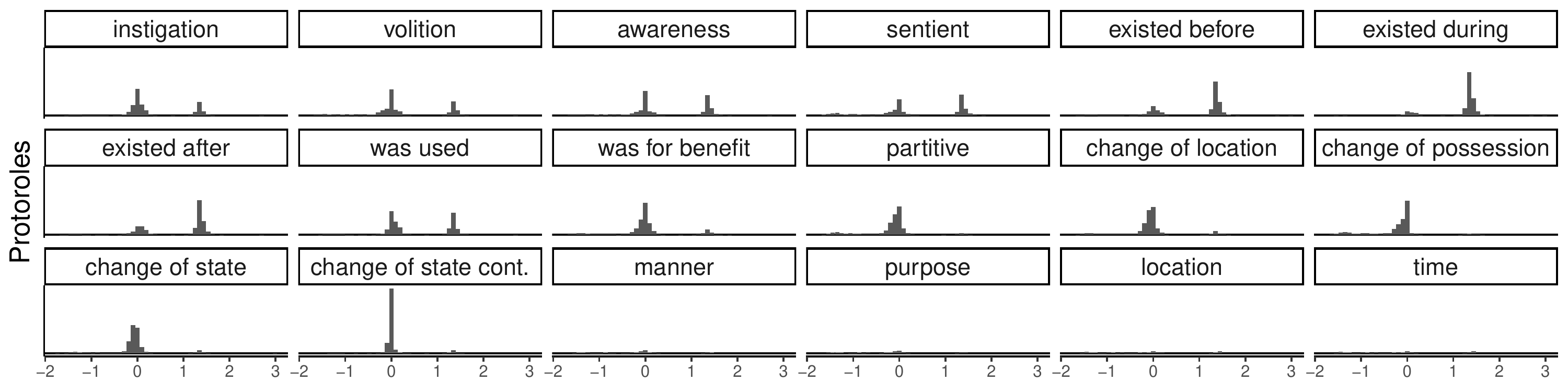}
    \vspace{-8mm}
    \caption{Distribution of attribute values in training and development sets. Only a subset of entity types are shown. Cases where the entity type attribute values default to the minimum are also excluded.}
    \label{fig:histogram}
\end{figure*}

There are currently four node type subspaces in UDS1.0: (i) factuality; (ii) genericity; (iii) time; and (iv) entity type. There is currently one edge type subspace: semantic proto-roles. For each attribute annotated on a particular node or edge, UDS1.0 provides two values: (i) the \textit{attribute value} itself (a real value) and (ii) a \textit{researcher confidence score} (a value on $[0, 1]$). The attribute value combines information about both (a) whether the attribute holds---in its sign---and (ii) the \textit{annotator confidence responses} made available with each dataset (in a form that depends on the particular dataset). The researcher confidence score quantifies our certainty that the attribute value accurately reflects all annotators' responses. This has the consequence that the more variable annotators responses are, the lower the researcher confidence score will be. 

Both the attribute value and the researcher confidence are derived from mixed effects models (MEMs). The goal in using these models is derive a single attribute value for each attribute on each node and edge that adjusts for idiosyncracies in how annotators use the particular instruments through which their annotations were collected, while simultaneously capturing variability across annotators' responses to the same item. Such adjustment is not possible with simpler methods---e.g. just taking the mean response across annotators.

Because each dataset uses a distinct annotation protocol with distinct annotation instruments---e.g. some datasets, such as the factuality dataset \cite{white_universal_2016,rudinger_neural_2018}, are collected using a binary instrument with ordinal confidence responses, while others, such as the semantic protoroles dataset \cite{reisinger_semantic_2015,white_universal_2016} use an ordinal instrument with binary confidence responses---the particular mixed model used for each differs (see \S\ref{sec:nodeattributes} and \S\ref{sec:edgeattributes} for details). But each model conforms to the same overarching principle: for each attribute and each node or edge that that attribute applies to, we assume (i) that there is some true, fixed real value for the node or edge on that attribute; (ii) that each annotator (drawn randomly from the annotator population) maps that attribute value to the response scale in a way specific to that annotator; and (iii) that each response by an annotator should be weighted by their reported confidence for that response (normalized to account for the fact that some annotators take different approaches to reporting confidence). Thus, we treat the attribute value as a fixed effect in a MEM (subsequently $z$-scored), the annotator response mappings as random effects, and the normalized annotator confidence score (always on $[0, 1]$) as a weight on the MEM's loss function. The distribution of these attribute value scores is shown in Figure \ref{fig:histogram}.

We derive our researcher confidence scores from these MEMs. Each MEM we use is probabilistic in the sense that the fixed attribute value and the random annotator mappings are optimized to maximize the likelihood of the observed responses by mapping these values through some \textit{link function}---e.g. for the datasets that use binary responses, we use a standard logistic as the link function. We compute the researcher confidence score for a particular attribute value of a particular node or edge as the mean of the likelihoods assigned to annotator responses for that attribute on that node or edge, weighted by the normalized annotator confidence response for that attribute on that node or edge. This has the effect that, if all annotators agree on a particular annotation with high (normalized) confidence, then the attribute value will be extreme and researcher confidence will be high; if half the annotators respond one way with high confidence and the other half responds another with high confidence, then researcher confidence will be low, since the value will be middling and will not assign particularly high likelihood to an particular response; and if half the annotators respond one way with high confidence and the other half responds another with low confidence, then researcher confidence will be middling. 

\subsubsection{Node attributes}
\label{sec:nodeattributes}

The four node type subspaces in UDS1.0---factuality, genericity, time, and entity type---are all derived from datasets collected by presenting annotators with a sentence containing a highlighted (possibly discontiguous) span corresponding to either a predicate (factuality, genericity, time) or an argument (genericity, entity type).

\paragraph{Factuality}

The UDS-Factuality dataset (v2.0) annotates predicates for whether or not the event or state that they refer to happened according to the author of the containing sentence \cite{white_universal_2016,rudinger_neural_2018}. Annotations are collected as binary responses along with an ordinal (1-5) confidence rating. 

\newcommand{\yb}{\mathbf{y}}

To normalize the ordinal confidence ratings across participants, we ridit score each participants' confidence ratings \cite{agresti_categorical_2014}. In ridit scoring, ordinal labels are normalized to (0, 1) using the empirical cumulative distribution function of the ratings given by each annotator. Specifically, for the responses $\yb_a$ given by annotator $a$:\footnote{see Govindarajan et al. 2019 for a recent use of such scoring in an NLP context, along with an intuitive explanation of its use.} 

\vspace{-5mm}

\[\text{ridit}_{\yb_a}\left(y_{ai}\right) = \text{ECDF}_{\yb_a}\left(y_{ai}-1\right) + 0.5 \times \text{ECDF}_{\yb_a}\left(y_{ai}\right)\]

\newcommand{\ub}{\mathbf{u}}

These normalized confidence ratings are entered as weights on the likelihood in a logistic mixed effects model. This model has real-valued fixed effects $\beta_i$ for each annotated predicate token $i$ and real-valued random intercepts $u_a$ for each annotator $a$. As is standard for mixed effects models, these parameters are optimized against a cross-entropy loss---in this case, a Bernoulli likelihood---with an additional term to enforce that random intercepts are normally distributed with mean $0$ and unknown variance $\sigma^2$.\footnote{The variance $\sigma^2$ is not optimized because that would result in driving it toward $\infty$; rather, it is estimated from $\ub$. This implies that the second term remains constant, and it is correct behavior, since this term is merely included to encode the assumption of random sampling over annotators by controlling the shape of the distribution of $\ub$.}

\vspace{-5mm}

\begin{align*}
\mathcal{L} = & \sum_i \sum_{a \in \alpha(i)} r_{ai}\log\mathrm{Bern}\left(y_{a\rho_a(i)}; p_{ai}\right)\\ 
&+ \sum_a \log\mathcal{N}\left(u_a; 0, \mathrm{Var}(\ub)\right)
\end{align*}

\vspace{-3mm}

where $\alpha(i)$ is the set of annotators that annotated predicate token $i$ for factuality, $\rho_a(i)$ is the index of the response to predicate $i$ within $y_{ai}$, $r_{ai} = \text{ridit}_{\yb_a}\left(y_{a\rho_a(i)}\right)$, and $p_{ai} = \mathrm{logit}^{-1}(\beta_i + u_a)$. We then take $\beta_i$ as the attribute value for predicate token $i$.

To derive the researcher confidence score $c_i$ for a predicate token $i$, we compute the mean of the probabilities of the annotations weighted by $r_{ai}$:

\vspace{-3mm}

\[c_i = \frac{\sum_{a \in \alpha(i)} r_{ai}\mathrm{Bern}\left(y_{a\rho_a(i)}; p_{ai}\right)}{\sum_{a \in \alpha(i)} r_{ai}}\]

\paragraph{Genericity}

The UDS-Genericity dataset (v1.0) annotates both predicates and arguments for a variety of attributes relevant to linguistic expression of generalization, including (i) whether or not a predicate refers to (a) a particular event or state (or some collection thereof); (b) a dynamic event; (c) a hypothetical situation; and (ii) whether or not an argument refers to (a) a particular thing or collection thereof; (b) a kind of thing; or (c) an abstract object \cite{govindarajan_decomposing_2019}. Like UDS-Factuality, annotations are collected as binary responses along with an ordinal (1-5) confidence rating, and so we use the same approach for constructing attribute values and research confidence scores used there.

\paragraph{Time}

The UDS-Time dataset (v1.0) annotates predicates for the likely duration of the event or state they refer to---whether it was \textit{instantaneous} or lasted \textit{seconds}, \textit{minutes}, \textit{hours}, \textit{days}, \textit{weeks}, \textit{months}, \textit{years}, \textit{decades}, \textit{centuries}, or \textit{forever}---along with an ordinal (1-5) confidence response. It also annotates pairs of predicates for the continuous temporal relation between the events or states the predicates in that pair refer to, along with an ordinal confidence response. We include only the duration annotations in UDS1.0 for reasons mentioned above.

We normalize the confidence ratings using ridit scoring, as for UDS-Factuality and UDS-Genericity. For normalizing the duration responses themselves, there are two reasonable options that are both generalizations of the approach taken for binary responses. The first would be to treat the duration responses as ordinal variables and induce a single duration value using an ordinal link logit model \cite{agresti_categorical_2014}. The second is to treat them as nominal variables and induce a real value for each duration using a multinomial logistic mixed model. 

We disprefer the first approach for two reasons. First, in mapping this response to a single real value, we lose information about the real world duration. The duration could be recovered by providing the ordinal model's binning of the real scale into duration values, but we take this indirection to be suboptimal. Second, we lose information about possible ambiguity in the duration leading to multimodal responses---e.g. being sick could be something that lasts for days or weeks, but it could also refer to a lifelong affliction, lasting decades. Ambiguity-driven multimodality is problematic for all our annotations---this is one reason why we include a researcher confidence score---but it is particularly problematic here in light of the first problem.  

As such, we implement the second approach, which yields a real-valued attribute corresponding to each duration. We derive this attribute from a multinomial logistic mixed effects model analogously to how we derive values for binary responses. In this case, the fixed effects $\bm{\beta}_i$ for each predicate token $i$ and the random effects $\ub_a$ for annotator $a$ are vectors of length equal to the number of duration responses.

\newcommand{\pb}{\mathbf{p}}
\newcommand{\Ub}{\mathbf{U}}

\vspace{-5mm}

\begin{align*}
\mathcal{L} = & \sum_i \sum_{a \in \alpha(i)} r_{ai}\log\mathrm{Cat}\left(y_{a\rho_a(i)}; \pb_{ai}\right)\\ 
&+ \sum_{a,k} \log\mathcal{N}\left(u_{ak}; 0, \mathrm{Var}(\ub_{\cdot k})\right)
\end{align*}

\vspace{-2mm}

where $\pb_{ai} = \mathrm{softmax}(\bm{\beta}_i + \ub_{a})$ and $\mathrm{Var}(\Ub)$ is a covariance matrix estimated from $\Ub$. We then take $\beta_{ik}$ as the attribute value for duration $k$ for predicate token $i$. We derive the corresponding research confidence score $c_{ik}$ analogously to what was done for binary responses.

\vspace{-3mm}

\[c_{ik} = \frac{\sum_{a \in \alpha(i)} r_{ai}\mathrm{Cat}\left(y_{a\rho_a(i)}; \pb_{ai}\right)}{\sum_{a \in \alpha(i)} r_{ai}}\]

\paragraph{Entity type}

The UDS-WordSense dataset (v1.0) annotates (the nominal heads of) arguments for the WordNet 3.0 \cite{miller_wordnet:_1995,fellbaum_wordnet_1998} senses that those (nominal heads of) arguments can have. For any particular argument, annotators were presented with all of the definitions of senses listed in WordNet for the head of that argument and asked to select all that were applicable using check boxes. After MEM-based normalization of the sense responses (described below), we extract entity types for these annotations by mapping the selected senses to their supersenses/lexicographer classes \cite{ciaramita_supersense_2003} and deriving a real-valued attribute value for each supersense from the normalized values associated with the senses that fall under it.

To normalize the sense responses, we use a logistic mixed effects model with real-valued fixed effects $\beta_{ik}$ for each annotated argument (head) token $i$ and potential sense $k$ and real-valued random intercepts $u_a$ for each annotator $a$:\footnote{Unlike the annotations from which we derive the other three node type subspaces, UDS-WordSense does not contain annotator confidence responses. We thus do not weight the likelihood of the MEM by normalized annotator confidence responses.}

\vspace{-5mm}

\begin{align*}
\mathcal{L} = & \sum_{i} \sum_{k \in \pi(i)} \sum_{a \in \alpha(i)} \log\mathrm{Bern}\left(y_{a\rho_a(i)k}; p_{aik}\right)\\ 
&+ \sum_a \log\mathcal{N}\left(u_a; 0, \mathrm{Var}(\ub)\right)
\end{align*}

\vspace{-3mm}

where $\pi(i)$ is the set of potential senses for argument (head) token $i$, and $p_{aik} = \mathrm{logit}^{-1}(\beta_{ik} + u_a)$. We then take $\beta_{ik}$ as the attribute value for predicate token $i$ and sense $k$.

To derive the researcher confidence score $c_{ik}$ for a predicate token $i$ and potential sense $k$, we compute the mean of the probabilities of the annotations:

\vspace{-3mm}

\[c_{ik} = \frac{\sum_{a \in \alpha(i)} \mathrm{Bern}\left(y_{a\rho_a(i)k}; p_{aik}\right)}{|\alpha(i)|}\]

We compute the attribute value $\gamma_{il}$ and research confidence $d_{ik}$ for each argument head token and supersense $l$ from $\beta_{ik}$ and $c_{ik}$ for all senses $k$ that fall under supersense $l$:

\[\gamma_{il} = \begin{cases}\max_{k \in \pi(i) \cap \psi(l)} \beta_{ik} & \pi(i) \cap \psi(l) \neq \emptyset\\\min_{i,k} \beta_{ik} & \text{otherwise} \end{cases}\]

where $\psi(l)$ is the set of senses that fall under supersense $l$. The research confidence is computed analogously:

\[d_{il} = \begin{cases}\max_{k \in \pi(i) \cap \psi(l)} c_{ik} & \pi(i) \cap \psi(l) \neq \emptyset\\1 & \text{otherwise} \end{cases}\]

We default to a confidence of $1$ here because if no sense of an argument (head) can fall under a particular supersense, then we have high confidence that the value should be low.

\subsubsection{Edge attributes}
\label{sec:edgeattributes}

The single node type subspaces in UDS1.0---the UDS-Protoroles (v2.x) dataset \cite{white_universal_2016}---is derived from a dataset collected by presenting annotators with a sentence containing two highlighted (possibly discontiguous) spans corresponding to a predicate and an argument. Annotators responded to 18 questions on an ordinal (1-5) scale, all starting with \textit{how likely or unlikely is it that...}\footnote{Questions 1-14 are modified versions of the questions used for the UDS-Protoroles (v1.0) dataset \cite{reisinger_semantic_2015} and were asked about arguments headed by the subject or direct object of the predicate's head. Questions 15-18 were asked about a distinct (and much smaller) set of arguments/adjuncts that were dependents of the predicate's head, but not subjects or direct objects. This is why the histograms for these attributes in Figure \ref{fig:histogram} are so low compared to the other questions.} 

\begin{enumerate}[itemsep=-3pt]
    \item \texttt{instigation}: ...\textsc{arg} caused the \textsc{pred} to happen?
    \item \texttt{volition}: ...\textsc{arg} chose to be involved in the \textsc{pred}?
    \item \texttt{awareness}: ...\textsc{arg} was/were aware of being involved in the \textsc{pred}?
    \item \texttt{sentient}: ...\textsc{arg} was/were sentient? 
    \item \texttt{change of location}: ...\textsc{arg} changed location during the \textsc{pred}?
    \item \texttt{existed before}: ...\textsc{arg} existed before the \textsc{pred} began?
    \item \texttt{existed during}: ...\textsc{arg} existed during the \textsc{pred}?
    \item \texttt{existed after}: ...\textsc{arg} existed after the \textsc{pred} stopped?
    \item \texttt{change of possession}: ...\textsc{arg} changed possession during the \textsc{pred}?
    \item \texttt{change of state}: ...\textsc{arg} was/were altered or somehow changed during or by the end of the \textsc{pred}?
    \item \texttt{change of state continuous}: ...the change in \textsc{arg} happened throughout the \textsc{pred}? (\textit{only shown if the answer to} \texttt{change of state} \textit{was 4 or 5})
    \item \texttt{was used}: ...\textsc{arg} was/were used in carrying out the \textsc{pred}?
    \item \texttt{was for benefit}: ...\textsc{pred} happened for the benefit of \textsc{arg}?
    \item \texttt{partitive}: ...\emph{only} a part or portion of \textsc{arg} was involved in the \textsc{pred}?
    \item \texttt{manner}: ...\textsc{arg} was the manner of the \textsc{pred}?
    \item \texttt{purpose}: ...\textsc{arg} was the purpose of the \textsc{pred}?
    \item \texttt{location}: ...\textsc{arg} was the location of the \textsc{pred}?
    \item \texttt{time}: ...\textsc{arg} was when the \textsc{pred} happened?
\end{enumerate}

UDS-Protoroles does not provides confidence annotations beyond whatever notion of confidence is part of giving the ordinal response to these questions; however, if the annotator responded with a 3 or less, an additional question was revealed asking whether the question was applicable. We normalize the ordinal response and applicability response separately and then combine the normalized ratings.

For the ordinal response, we use an ordinal link logit mixed effects model \cite{agresti_categorical_2014} with real-valued fixed effects $\beta_{ik}$ for each annotated predicate-argument pair $i$ and each property $k$ and real-valued random intercepts $\ub_a$ for each annotator $a$. This sort of model, which has been used in prior work on the semantic protoroles (v1.0) dataset \cite{white_semantic_2017}, is a straightforward generalization of logistic regression to more than two labels. The random intercepts $\ub_a$ are a vector with four monotonically increasing elements that separates the real values into five bins corresponding to the five ordinal responses. These intercepts (or \textit{cutpoints}) are used to define a cumulative categorical probability distribution, from which the probability mass function for said distribution can be reconstructed.

\newcommand{\prob}{\mathbb{P}}
\vspace{-5mm}
\[\prob(y_{a\rho_a(i)k} \leq l) = \begin{cases}\text{logit}^{-1}(\beta_{ik} - u_{al}) & \text{if } l \in \{1, ..., 4\}\\
1 & \text{if } l = 5\\
0 & \text{otherwise}\end{cases}\]

\vspace{-2mm}

The probability $p_{aikl}$ of an ordinal response $l$ for a predicate-argument pair $i$ on property $k$ by annotator $a$ is thus defined as:

\vspace{-4mm}

\[p_{aikl} = \prob(y_{a\rho_a(i)k} \leq l) - \prob(y_{a\rho_a(i)k} \leq l-1)\]

The likelihood is then a simple categorical likelihood. The loss term for the random effects places a distribution with strictly positive support---here, an exponential---on the distance between the random intercepts.

\vspace{-5mm}

\begin{align*}
\mathcal{L} = & \sum_i \sum_{a \in \alpha(i)} \log\mathrm{Cat}\left(y_{a\rho_a(i)k}; \pb_{aik}\right)\\ 
&+ \sum_{a,l} \log\text{Exp}\left(u_{al} - u_{a(l-1)}; \frac{1}{\mathrm{Var}(\ub_{\cdot l} - \ub_{\cdot (l - 1)})}\right)
\end{align*}

\vspace{-2mm}

We derive the corresponding researcher confidence score $c_{ik}$ analogously to the node type subspaces.

\vspace{-3mm}

\[c_{ik} = \frac{\sum_{a \in \alpha(i)} \mathrm{Cat}\left(y_{a\rho_a(i)k}; \pb_{aik}\right)}{|\alpha(i)|}\]

The applicability ratings are normalized using a logisitic mixed effects model to yield fixed effects $\delta_{ik}$ for each predicate argument pair $i$ and property $k$. The final normalized score for a predicate argument pair $i$ and property $k$ is then computed as $\text{logit}^{-1}(\delta_{ik})\beta_{ik}$. This pulls properties that are not applicable for a particular pair toward zero.

\section{Toolkit}

The Decomp toolkit (v0.1) is a Python 3 package that provides utilities for: 

\begin{enumerate}[itemsep=-3pt]
    \item reading the the UDS dataset from the underlying treebank and annotations or directly from its native JSON format, including facilities for quickly adding user-defined annotations to the graphs
    \item serializing UDS graphs to many common formats, such as Notation3, N-Triples, turtle, and JSON-LD, as well as any other format supported by NetworkX
    \item querying both the syntactic and semantic subgraphs of UDS (as well as pointers between them) using SPARQL 1.1 queries
\end{enumerate}

This last feature is particularly useful for quickly and easily searching for sentences based on complex syntactic and semantic constraints. These queries can be relatively simple. For example, if one were interested in extracting only predicates referring to events that likely happened and likely lasted for minutes:
\vspace{-1mm}
{\footnotesize
\begin{verbatim}
SELECT ?pred
WHERE { ?pred <domain> <semantics> ;
              <type> <predicate> ;
              <factual> ?factual ;
              <dur-minutes> ?duration
              FILTER ( ?factual > 0 && 
                       ?duration > 0 
                      )
      }
\end{verbatim}
}

But they can also be arbitrarily sophisticated.  For instance, if one were interested in extracting all predicate-argument edges that (i) touch a predicate referring to an event that is likely spatiotemporally delimited; and (ii) touch at least one argument that refers to a spatiotemporally delimited participant that was volitional in the event:
\vspace{-1mm}
{\footnotesize
\begin{verbatim}
SELECT ?edge
WHERE { ?pred ?edge ?arg ;
              <domain> <semantics> ;
              <type>   <predicate> ;
              <pred-particular> ?ppart
              FILTER ( ?ppart > 0 ) .
        ?arg  <domain> <semantics> ;
              <type>   <argument>  ;
              <arg-particular> ?apart
              FILTER ( ?apart > 0 ) .
        { ?edge <volition> ?volition
          FILTER ( ?volition > 0 )
        } UNION
        { ?edge <sentient> ?sentient
          FILTER ( ?sentient > 0 )
        }
      }
\end{verbatim}
}

Further, syntactic and semantic constraints can be mixed. For instance, if one were interested in all copular predicates with at least one argument that refers to a spatiotemporally delimited participant that was sentient in the event referred to by the predicate: 

\vspace{-1mm}

{\footnotesize
\begin{verbatim}
SELECT ?pred
WHERE { ?pred ?semedge ?arg ;
              <domain> <semantics> ;
              <type>   <predicate> .
        ?arg  <domain> <semantics> ;
              <type>   <argument>  ;
              <arg-particular> ?apart
              FILTER ( ?apart > 0 ) .
        ?semedge <sentient> ?sentient
           FILTER ( ?sentient > 0 ) .
        ?pred ?instedge ?head .
        ?instedge <domain> <interface> ;
                  <type>   <head> .
        ?head ?synedge ?syndep .
        ?syndep <deprel> ?relation
           FILTER ( ?relation = "cop" ) .
      }
\end{verbatim}
}

\section{Conclusion}

We presented the Universal Decompositional Semantics dataset (v1.0), which is bundled with the Decomp toolkit (v0.1) and discussed how we construct the Universal Decompositional Semantics (UDS) dataset (v1.0) by unifying five high-quality, decompositional semantics-aligned annotation sets within a single semantic graph specification based on the predicative patterns produced by the PredPatt tool. We also presented the Decomp toolkit (v0.1), which provides a suite of Python 3 tools for querying Universal Decompositional Semantics graphs using SPARQL 1.1. We believe these resources will be helpful to (i) those wishing to pursue corpus linguistic studies on the existing annotations; and (ii) those pursuing broad-coverage semantic parsing algorithms, where the structural distinctions in UDS, as compared to previously existing semantic annotated corpora, may offer unique challenges.

\section*{Acknowledgements}

This research
was supported by the University of Rochester,
JHU HLTCOE, the National Science Foundation (BCS-1748969/BCS-1749025), DARPA AIDA, DARPA KAIROS, and IARPA BETTER. The U.S.
Government is authorized to reproduce and distribute reprints for Governmental purposes. The
views and conclusions contained in this publication are those of the authors and should not be
interpreted as representing official policies or endorsements of DARPA or the U.S. Government.

\section*{References}
\label{main:ref}

\bibliographystyle{lrec}
\bibliography{references}


\end{document}